\documentclass{article}
\pdfoutput=1

\usepackage[utf8]{inputenc} 
\usepackage[T1]{fontenc}    
\usepackage{url}            
\usepackage{booktabs}       
\usepackage{amsfonts}       
\usepackage{nicefrac}       
\usepackage{microtype}      
\usepackage{graphicx}
\usepackage{subcaption}
\usepackage{booktabs} 
\usepackage[square,sort,comma,numbers]{natbib}

\usepackage[final]{nips_2018}

\usepackage{here}
\usepackage{graphicx}
\usepackage{caption}
\usepackage[ruled,noend]{algorithm2e}
\usepackage{amsmath,amssymb,amsfonts,amsbsy,amsfonts,latexsym}
\usepackage{amsthm}
\usepackage{multirow}
\usepackage{makecell}
\usepackage[labelfont=bf,textfont=it,belowskip=0pt,aboveskip=5pt,tableposition=top]{caption}
\usepackage{xcolor}
\usepackage{colortbl}

\usepackage{wrapfig}

\definecolor{colorA}{RGB}{189,201,225}
\definecolor{colorB}{RGB}{103,169,207}
\definecolor{colorC}{RGB}{ 28,144,153}
\definecolor{colorD}{RGB}{  1,108, 89}

\newcolumntype{R}{>{\columncolor{gray!40}}r}
\newcolumntype{L}{>{\columncolor{gray!40}}l}
\newcolumntype{C}{>{\columncolor{gray!40}}c}

\usepackage{tabularx,colortbl,xcolor}
\usepackage{multirow}
\usepackage[normalem]{ulem}
\useunder{\uline}{\ul}{}

\newcommand\Ga{}

\newcommand\Gc{\rowcolor{gray!30}}

\usepackage{enumitem}

\usepackage{xparse}

\graphicspath{{figs}}
\DeclareGraphicsExtensions{.pdf,.png}
\usepackage{pgfplots, pgfplotstable}

\def\0{{\bf 0}}
\def\1{{\bf 1}}

\NewDocumentCommand{\var}{O{s} m O{}}{%
  \ensuremath{#1_{#2}^{#3}}
}
\usepackage{siunitx}

\definecolor{light-gray}{gray}{0.80}

\renewcommand\paragraph{\subsubsection*}

\newcommand\fref{Fig.~\ref}
\newcommand\tref{Table~\ref}
\SetKwInOut{Parameter}{parameter}

\def\0{{\bf 0}}

\newcommand\atsign{@}

\title{Parameter Re-Initialization through Cyclical Batch Size Schedules}

\newcommand*\samethanks[1][\value{footnote}]{\footnotemark[#1]}

\author{Norman Mu\thanks{Equal contribution}~~~Zhewei Yao\samethanks~~~Amir Gholami~~~Kurt Keutzer~~~Michael W. Mahoney\\
\{thenorm, zheweiy, amirgh, keutzer and mahoneymw\}\atsign berkeley.edu
}

\begin{document}

\maketitle
\begin{abstract}
Optimal parameter initialization remains a crucial problem for neural network training.
A poor weight initialization may take longer to train and/or converge to sub-optimal solutions. 
Here, we propose a method of weight re-initialization by  
repeated annealing and injection of noise in the training process. We implement this through a cyclical batch size schedule motivated by a Bayesian perspective of neural network training.
We evaluate our methods through extensive experiments on tasks in language modeling, natural language inference, and image classification.
We demonstrate the ability of our method to improve language modeling performance by up to 7.91 perplexity and reduce training iterations by up to $61\%$, in addition to its flexibility in enabling snapshot ensembling and use with adversarial training.
\end{abstract}
\section{Introduction}\label{sec:intro}
Despite many promising empirical results at using stochastic optimization methods to train highly non-convex modern deep neural networks, we still lack theoretically robust practical methods which are able to escape saddle points and/or sub-optimal local minima and converge to parameters that retain high testing performance.  
This lack of understanding leads to practical training challenges.

Stochastic Gradient Descent (SGD) is currently the de-facto optimization method for training
deep neural networks (DNNs). 
Through 
extensive hyper-parameter tuning, SGD 
can avoid poor local optima and achieve good generalization ability.
One important hyper-parameter that can
significantly affect SGD performance is the weight initialization. For instance, initializing the 
weights to all zeros or all ones leads to extremely poor performance~\cite{xu2017second}.
Different approaches have been proposed for weight initialization such
as Xavier, MSRA, Ortho, LSUV~\cite{glorot2010understanding,he2015delving,saxe2013exact,mishkin2015all}.
These are mostly agnostic to the model architecture and the specific learning task.

Our work explores the idea of adapting the weight initialization to the optimization dynamics of the specific learning task at hand.
From the Bayesian perspective, improved weight initialization can be viewed
as starting with a better prior, which leads to a more accurate
posterior and thus better generalization ability. 
This problem has been explored extensively in Bayesian optimization. For example, in the seminal works~\cite{hu2017adaptive,roberts2009examples}, an adaptive
prior is implemented via Markov Chain Monte Carlo (MCMC) methods. 
Motivated by these ideas, we incorporate an ``adaptive initialization'' for neural network training (see section~\ref{sec:methods} for details), where we use cyclical batch size schedules to control the noise (or temperature) of SGD.
As argued in~\cite{smith2018bayesian}, both learning rate and batch size can be used to control the noise of SGD but the latter
has an advantage in that it allows more parallelization opportunity~\cite{gholami2017integrated}.
The idea of using batch size to control the noise in a simple cyclical schedule was recently proposed in~\cite{jastrzkebski2017three}.
Here, we build upon this work by studying different cyclical annealing strategies for a wide
range of problems. Additionally, we discuss how this can be combined with a new adversarial regularization scheme recently proposed in~\cite{yao2018large}, as well as prior work~\citep{huang2017snapshot} in order to obtain ensembles of models at no
additional cost.
In summary, our contributions are as follows:

\begin{itemize}
    \item We explore different cyclical batch size (CBS) schedules for training neural networks
    inspired by Bayesian statistics, particularly adaptive MCMC methods. The CBS schedule leads to multiple perplexity improvement (up to 7.91) in language modeling and minor improvements in natural language inference and image classification. 
    Furthermore, we show that CBS schedule can alleviate problems with overfitting and sub-optimal parameter initialization.
     
     \item Additionally, CBS schedules require up to $3\times$ fewer SGD iterations due to larger batch sizes, which allows for
     more parallelization opportunity. This reflects the benefit of cycling the batch size instead
     of the learning rate as in prior work~\citep{smith2017don, huang2017snapshot}
     
     \item We showcase the flexibility of CBS schedules for use with additional techniques. We propose a simple but effective ensembling method that combines models saved during different
     cycles at no additional training cost. 
     In addition, we show that CBS schedule can be combined with other approaches such as the recently proposed adversarial regularization~\cite{yao2018large} to yield further classification accuracy improvement of $0.26\%$.

\end{itemize}

\paragraph{Related Work} 
\cite{glorot2010understanding} introduced Xavier initialization, which keeps the variance of
input and output of all layers within a similar range in order to prevent vanishing or 
exploding values in both the forward and backward passes. Building off this 
idea,~\cite{he2015delving} explored a new strategy known as MSRA to keep the 
variance constant for all convolutional layers. \cite{saxe2013exact} proposed an 
orthogonal initialization (Ortho) to achieve faster convergence, 
and more recently,~\cite{mishkin2015all} combined ideas from previous work and showed 
that a unit variance orthogonal initialization is beneficial for deep models.

\citep{jastrzkebski2017three,smith2018bayesian,devarakonda2017adabatch} show that the noise of SGD is controlled by the ratio of learning rate to batch size. The authors argued that
the SGD algorithm can be derived through Euler-Maruyama discretization of a Stochastic Differential Equation (SDE).
The SDE dynamics are  governed by a "noise scale" $g \approx \epsilon N/B$ for $\epsilon$ the learning rate, $N$ 
the training dataset size, and $B$ the batch size. They conclude that a higher noise scale 
prevents SGD from settling into sharper minima. This result supports a prior empirical 
observation~\citep{krizhevsky2014one} that under certain mild assumptions such as $N \gg B$,
the effect of dividing the learning rate by a constant factor is equivalent to that of 
multiplying the batch size by the same constant factor.
In related work, \citep{smith2017don} applied this understanding and used batch size
as a knob to control the noise, and empirically showed that the baseline performance could be matched. \cite{yao2018large} further explored how to use second-order information and adversarial training to control the noise for training large batch size.
\cite{martin2017rethinking,martin2018implicit} showed using a statistical mechanics argument that many other hyper-parameters in neural network training, e.g. data quality, can also act as temperature knobs.
\section{Methods}\label{sec:methods}
The goal of neural network optimization is to solve an empirical risk minimization, with a loss function of the form:

\begin{equation}\label{eqn:basic_problem}
L(\theta) = \frac{1}{N} \sum_{i=1}^{N} l(x_i, \theta),
\end{equation}
where $\theta$ is the model parameters, $X$ is the training
dataset and $l(x, \theta)$ is the loss function. Here $N=|X|$ is the cardinality of the training set. In SGD, a mini-batch, $B\subset\{1,2,..., N\}$ is used to compute an (unbiased) gradient,
i.e., $g_{t} = \frac{1}{|B|} \sum_{x \in B} \nabla_\theta l(x, \theta_t)$,
and this is typically used to optimize \eqref{eqn:basic_problem} in the form:
\begin{equation}\label{eqn:sgd}
\theta_{t+1} = \theta_t - \eta_t g_t,
\end{equation}
where $\eta_t$ is the learning rate (step size) at iteration $t$, and commonly annealed during training.

By Bayes' Theorem, given the input data, $X$, a prior distribution on the model 
parameters, $P(\theta)$, and a likelihood function, $P(X|\theta)$, the posterior distribution,  $P(\theta | X)$, is:
\begin{equation}
    P(\theta | X) \propto P(\theta)P(X|\theta).
\end{equation}

From this Bayesian perspective, the goal of the neural network training is to find the Maximum-A-Posteriori (MAP) point for a given prior distribution. Note that in this context weight initialization and prior distribution are similar, that is a better prior distribution would lead to more informative posterior.
In general, it may be difficult to design a better prior given only data and a model architecture. Additionally,
the high dimensionality of the NN's parameter space renders various approaches such as adaptive priors intractable (e.g. adaptive MCMC algorithms~\cite{roberts2009examples,hu2017adaptive}).
Hence, we look into an adaptive weight ``re-initialization'' strategy. We start with an input
prior (weight initialization) and compute an approximate MAP point by annealing the noise in SGD. 
Once we compute the MAP point, we use it as a new 
initialization of the neural network weights, and restart the noise annealing 
schedule. We then iteratively repeat this process through the training process.

\begin{figure}[!htbp]
 \centering
\includegraphics[width=.4\textwidth]{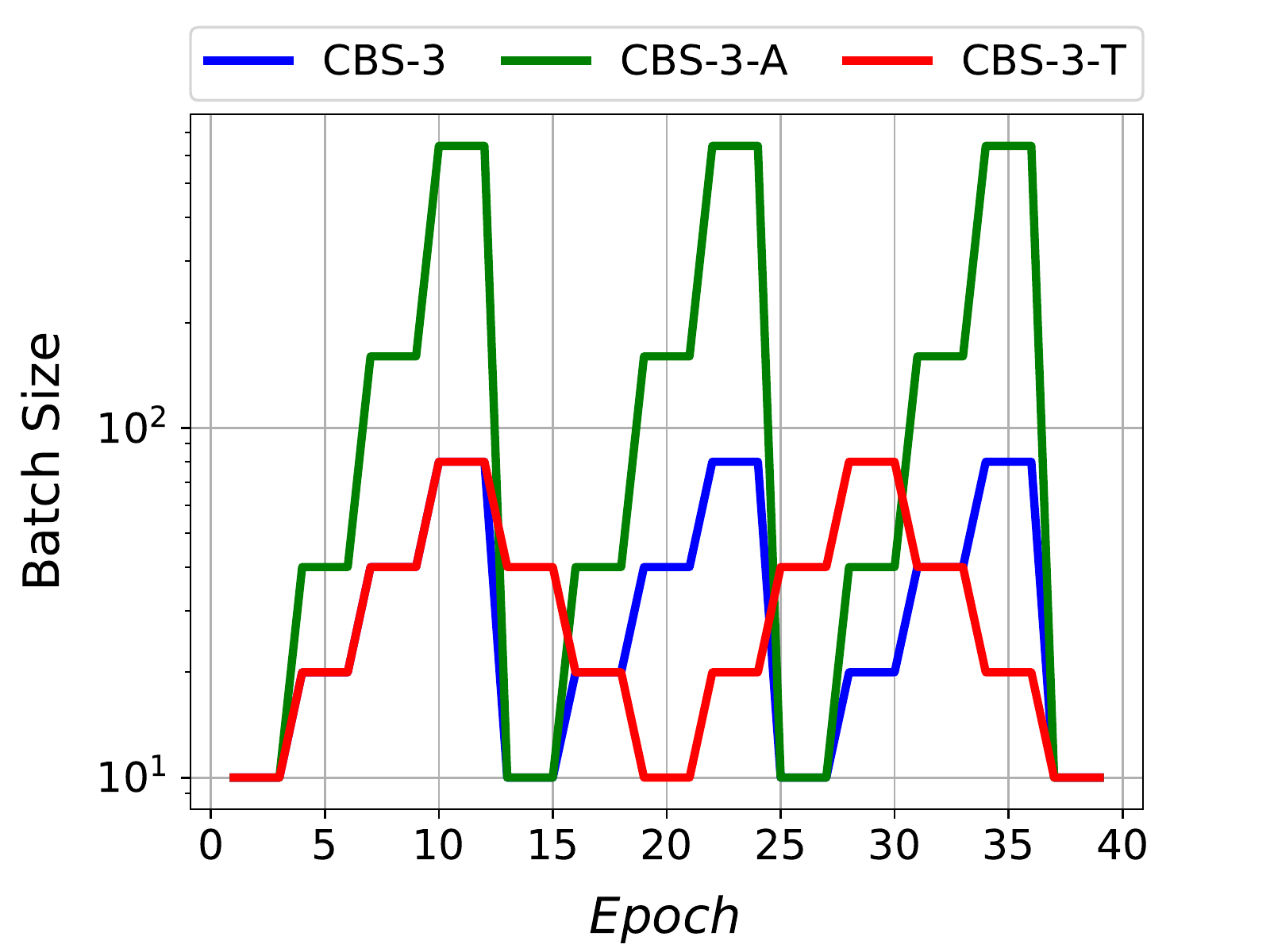}
\includegraphics[width=.4\textwidth]{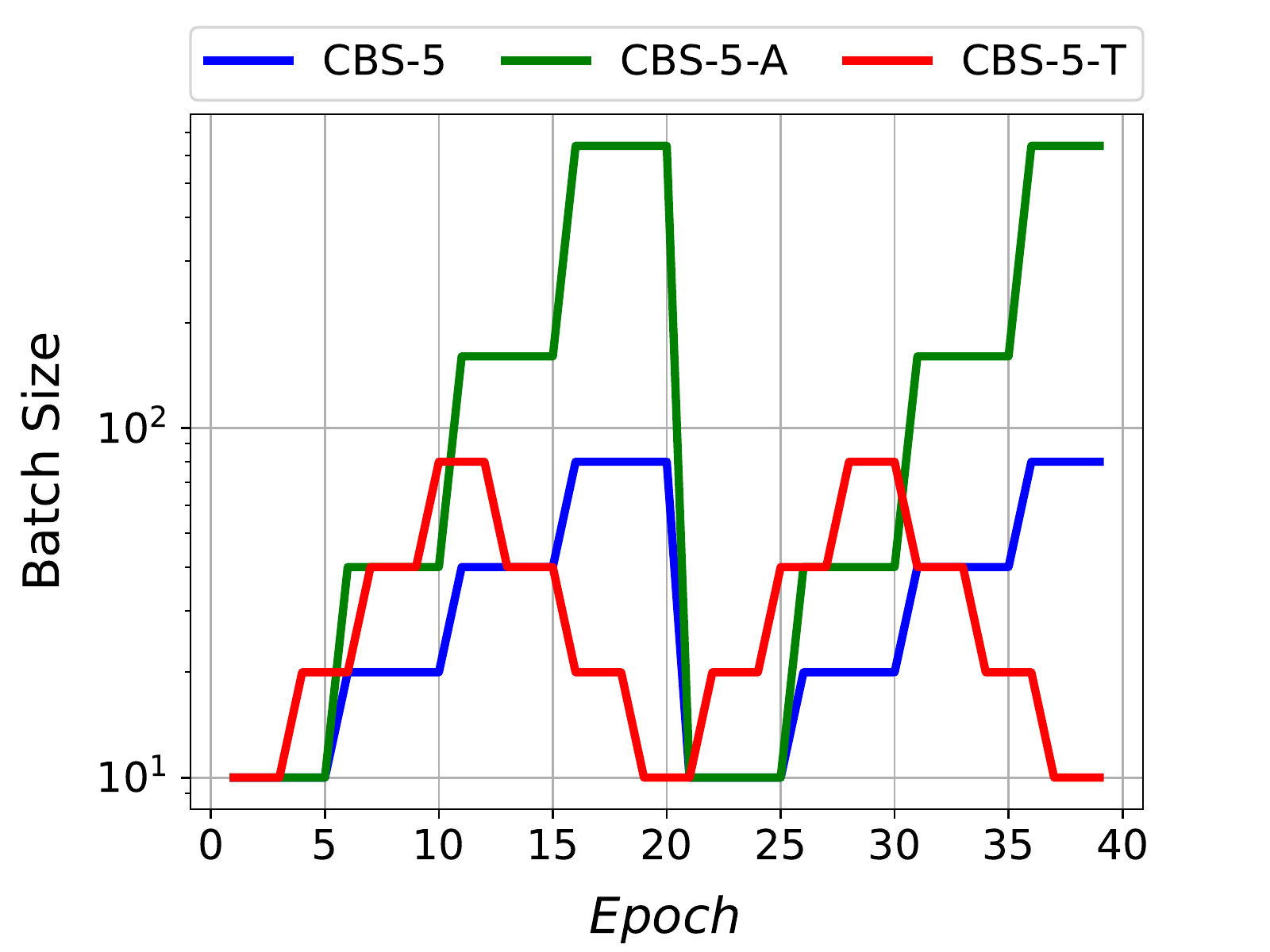}
 \caption{\footnotesize Illustration of 6 different CBS schedules, with initial batch size of 10; see Appendix~\ref{sec:training_outline} for details.} 
 \label{fig:cbs_cartoon}
\end{figure}

One approach to controlling the level of noise in SGD is via the learning rate, which is the approach used 
in~\cite{huang2017snapshot,smith2017cyclical}. However, as
discussed in~\cite{jastrzkebski2017three,smith2017don,devarakonda2017adabatch}, the batch size can also be used to control SGD noise. The motivation
for this is that larger batch sizes allow for parallel execution
which can accelerate training. 
We implement weight re-initialization through cyclical batch size schedules. 
The SGD training process is divided into one or more cycles, and in single cycle we gradually increase the batch size to decrease noise. As the noise level of SGD is annealed, $\theta$ will approaches a 
 local minima i.e., an approximate MAP point of $P(\theta| X)$.
Then at the beginning of the subsequent cycle we drop the batch size back down to the initial value, which increases the noise in SGD and "re-initializes" the neural network parameters using the previous estimate. Several CBS schedules are shown in~\fref{fig:cbs_cartoon}.

\section{Results}
We perform a variety of experiments across different tasks and neural network architectures in natural language processing as well as image classification. We report our experimental findings on language tasks in section~\ref{sec:NLP}, and image classification
in section~\ref{sec:image_class}. We illustrate that CBS schedules can alleviate sub-optimal initialization in section~\ref{sec:bad_init}. We follow the baseline training method for each task (for details please see Appendix~\ref{sec:training_outline}). 
Alongside testing/validation performance, we also report the number of training iterations (lower values are preferred).

\subsection{Language Results}\label{sec:NLP}

Language modeling is a challenging problem due to the complex and long-range interactions between distant words~\cite{merity2016pointer}.
One hope is that large/deep models might be able to 
capture these complex interactions, but large models easily overfit on these tasks and exhibit large gaps between training set and testing set performance. 
CBS schedules effectively help us avoid overfitting, and in addition snapshot ensembling enables even greater performance.

\begin{table}[!htbp]
\caption{\footnotesize Testing perplexity and number of parameter updates of L1 and L2 models on Penn 
Tree Bank (PTB) and WikiText~2 (WT2) datasets. The best perplexity and lowest number of 
updates are \textbf{bolded}. }
\label{tab:lm-results}
\centering
\begin{tabular}{lcc|cc|cc|cc} \toprule
  &\multicolumn{2}{c}{L1 on PTB}       &\multicolumn{2}{c}{L1 on WT2}    &\multicolumn{2}{c}{L2 on PTB} &\multicolumn{2}{c}{L2 on WT2}\\              
\midrule
Schedule    & {Per.}    & {\# Iters}    & {Per.}    & {\# Iters}    & {Per.}     & {\# Iters}   & {Per.}    & {\# Iters}      \\
\midrule
\Gc	BL\footnotemark	&	83.13	&	52k	&	96.41	&	116k	&	79.34	&	73k	&	99.69	&	164k	\\
\midrule
\Ga	CBS-10	&	80.49	&	49k	&	94.93	&	111k	&	79.37	&	83k	&	95.43	&	187k	\\
\Gc	CBS-5	&	80.78	&	49k	&	\textbf{94.31}	&	111k	&	78.61	&	73k	&	94.32	&	164k	\\
\Ga	CBS-1	&	81.56	&	49k	&	94.52	&	111k	&	77.56	&	69k	&	\textbf{91.78}	&	156k	\\
\midrule
\Gc	CBS-10-A	&	\textbf{80.28}	&	35k	&	95.91	&	79k	&	81.47	&	65k	&	95.28	&	146k	\\
\Ga	CBS-5-A	&	82.03	&	35k	&	95.23	&	79k	&	79.48	&	53k	&	93.63	&	118k	\\
\Gc	CBS-1-A	&	84.41	&	35k	&	95.66	&	79k	&	81.32	&	49k	&	93.19	&	111k	\\
\midrule
\Ga	CBS-10-T	&	80.49	&	49k	&	94.93	&	111k	&	79.42	&	83k	&	94.39	&	187k	\\
\Gc	CBS-5-T	&	80.94	&	53k	&	94.9	&	120k	&	78.95	&	63k	&	94.68	&	142k	\\
\Ga	CBS-1-T	&	81.82	&	46k	&	95.38	&	104k	&	\textbf{77.39}	&	65k	&	93.78	&	147k	\\
\bottomrule 
\end{tabular}
\end{table}

We evaluate a large variety of CBS schedules to positive results as shown in~\tref{tab:lm-results}. 
Results are measured in perplexity, a standard figure of merit for evaluating the quality of language models by measuring its prediction of the empirical distribution of words (lower perplexity value is better). 
As we can see, the best performing CBS schedules result in significant improvements in
perplexity (up to 7.91) over the baseline schedules and also offer reductions in the number of SGD training iterations (up to $33\%$). For example, CBS schedules achieve improvement of 7.91 perplexity improvement on WikiText~2 via CBS-1-T and reduce the SGD iterations from 164k to 111k via the CBS-1-A schedule.
Notice that almost all CBS schedules outperform the baseline schedule.
\begin{figure}[!htbp]
  \centering
\includegraphics[width=.4\textwidth]{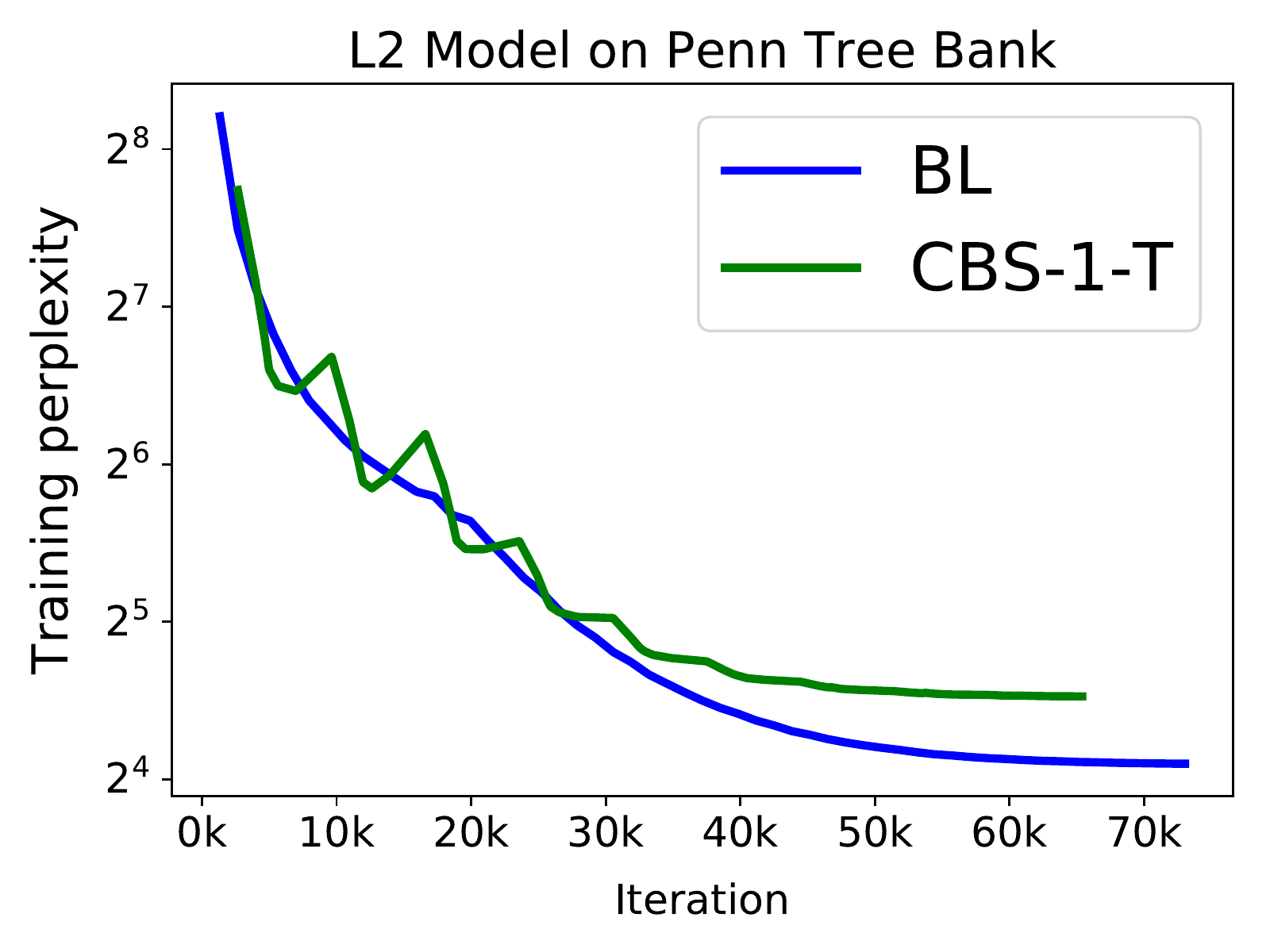}
\includegraphics[width=.4\textwidth]{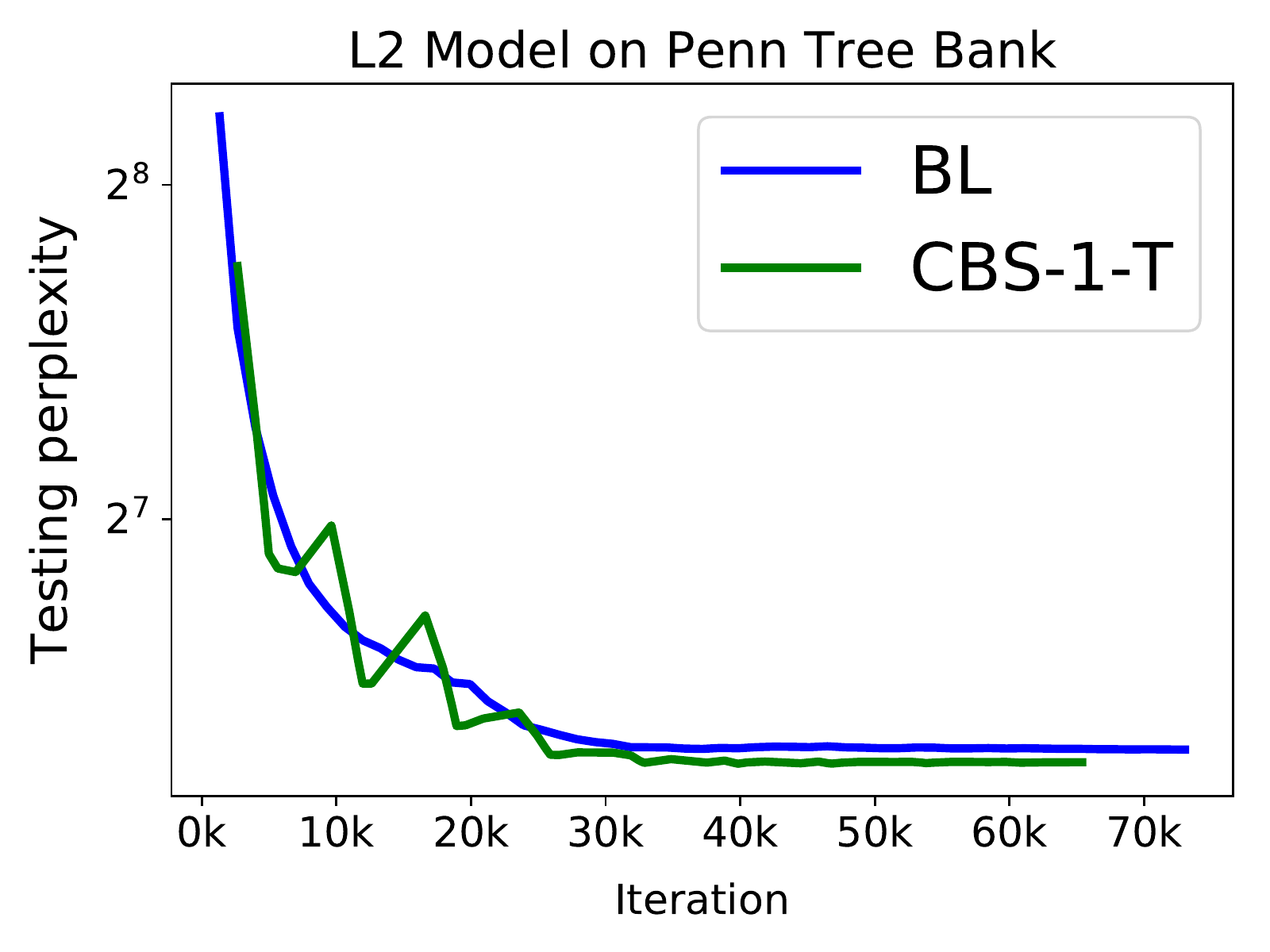}
  \caption{\footnotesize Training (left) and testing (right) perplexity as a function of iterations for the L2 model on PTB.}
  \label{fig:l2_ptb}
\end{figure}

\fref{fig:l2_ptb} shows the training and testing perplexity of the L2 model on PTB and WikiTest~2 as trained via the baseline schedule 
along with our best CBS schedule (from \tref{tab:lm-results}). Notice the cyclical spikes in training and testing perplexity. The peaks occur during  decreases in batch size, i.e., increases in noise scale, which could help to escape sub-optimal local minima, and the troughs occur during increases in batch size, i.e., decreases with noise scale.

In order to support our claim that CBS schedules are especially useful for counteracting overfitting, we conducted additional language modeling experiments on models L1', L2' with PTB and WT2 which use significantly lower dropout (0.2 and 0.3) than the original L1, L2 models (0.5 and 0.65). Because these models heavily overfit the training data, we report both the final testing perplexity as well as the best testing perplexity achieve during training. 
As seen in \tref{tab:lm-results-overfit} (in Appendix~\ref{sxn:app:additional results}), with L2' CBS yields improvements of a staggering 60.3 on final testing perplexity and 36.2 on best testing perplexity. CBS yields smaller improvements on L1' of 26.0 and 25.3, which are still much larger than the improvement achieved by CBS on L1 and L2.

As mentioned above the goal of every cycle is to get an approximate MAP point. A very interesting idea
proposed in~\citep{huang2017snapshot} is to ensemble these MAP points by saving snapshots of the model
at the end of every cycle. We follow that strategy with the only difference that we use a batch size cycle
instead of cyclical learning rate proposed in~\citep{huang2017snapshot} due to higher parallelization opportunities for the former.
We perform experiments on snapshot ensembling with the L2 model with the respective best performing CBS schedules on PTB and WikiText~2 (CBS-1-T and CBS-1), as well as the fixed batch size baseline.
The CBS ensembles on PTB and WikiText~2 result in test set perplexity of 76.14 and 88.47, outperforming baseline ensembles on both datasets (76.52, 89.99 respectively) and CBS single models (77.39, 91.78 respectively).

To further explore the properties of cyclical batch size schedules, we also evaluate these schedules on natural language inference tasks, as shown in~\tref{tab:nli-results}. In our experiments, CBS schedules do not yield large performance improvements on models like E1 which exhibit smaller disparities between training and testing performance.
This is in line with our limitation in that CBS is more effective for models which tend to overfit. On the other hand, we see a large reduction in training 
iterations by up to 62\% which is due to higher effective batch size used in CBS than baseline.

\footnotetext{\citep{zaremba2014recurrent} reports testing perplexity of 82.7 and 78.4 
for L1 and L2 respectively on PTB, which we could not reproduce. The best perplexity and lowest number of updates are \textbf{bolded}.}

\begin{wraptable}{r}{7cm}
\caption{\footnotesize Validation accuracy and number of parameter updates of E1 on MultiNLI and SNLI 
datasets. The best accuracy and lowest number of updates are \textbf{bolded}. }
\label{tab:nli-results}
\centering
\begin{tabular}{lcc|cc} \toprule
  &\multicolumn{2}{c}{MultiNLI}       &\multicolumn{2}{c}{SNLI}    \\              
\midrule
Strategy    & {Acc.}    & {\# Iters}    & {Acc.}    & {\# Iters}    \\
\midrule
\Gc	BL	    &	72.87	&	123k	&	\textbf{86.86}	&	172k	\\
\Ga	CBS-1	&	\textbf{73.17}	&	64k	&	86.73   &   90k   \\
\Gc	CBS-2	&	73.07	&	71k	&	86.56   &   99k	\\
\Ga CBS-1-A &   72.23   & \textbf{48k}     & 86.26     & \textbf{67k}     \\
\Gc CBS-2-A &   72.04   & 57k     & 85.83     & 80k     \\
\bottomrule 
\end{tabular}
\end{wraptable}

\subsection{Image Classification Results}\label{sec:image_class}
We also test our CBS schedules on Cifar-10 and ImageNet. Table.~\ref{tab:cbs_cifar10} reports the testing accuracy and the number of training iterations for different models on Cifar-10. We see that the CBS schedules match baseline performance, but the number of training iterations used in CBS schedules is up to $2\times$ fewer. 

\begin{figure}[!htbp]
  \centering
\includegraphics[width=.4\textwidth]{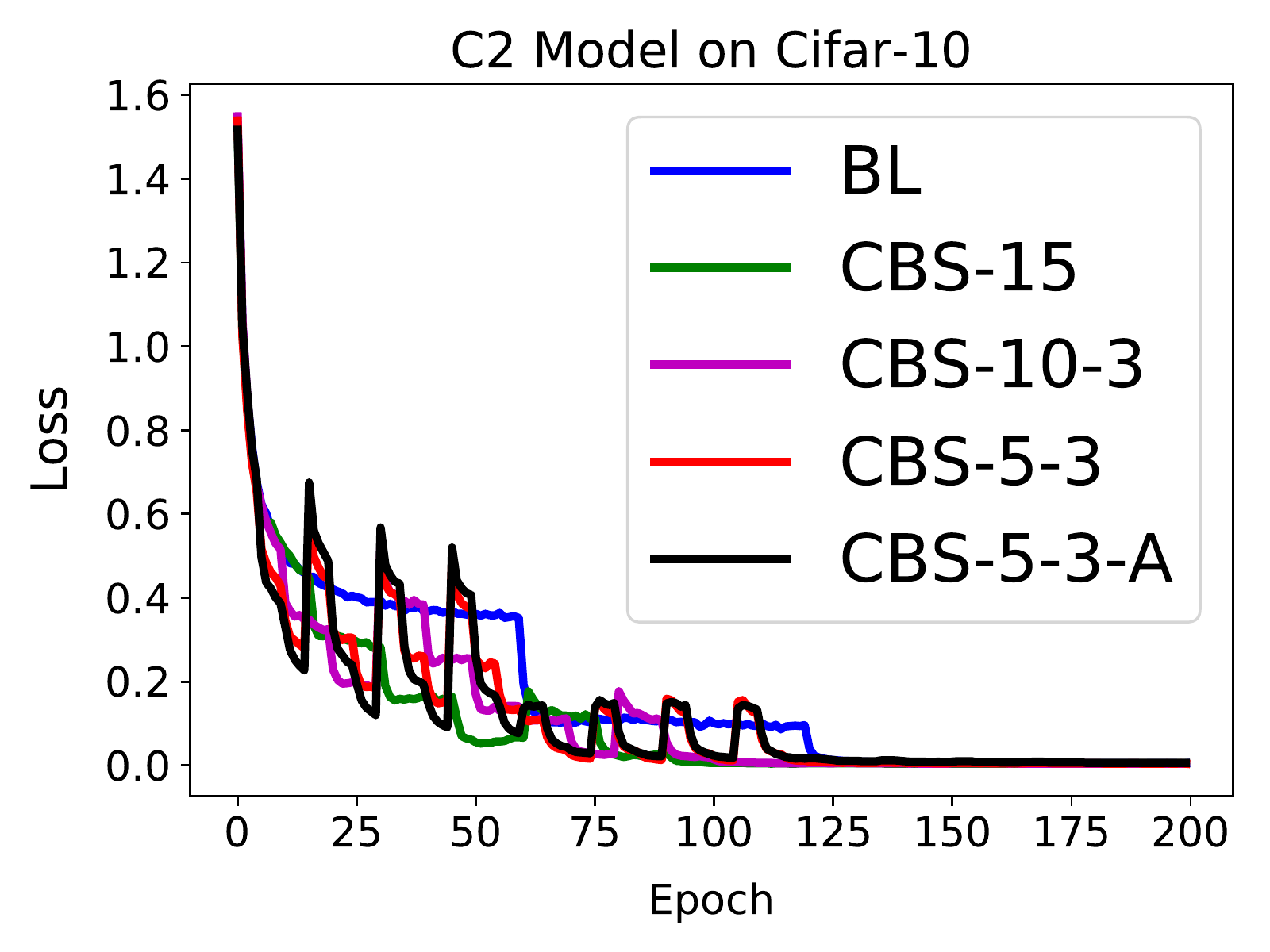}
\includegraphics[width=.4\textwidth]{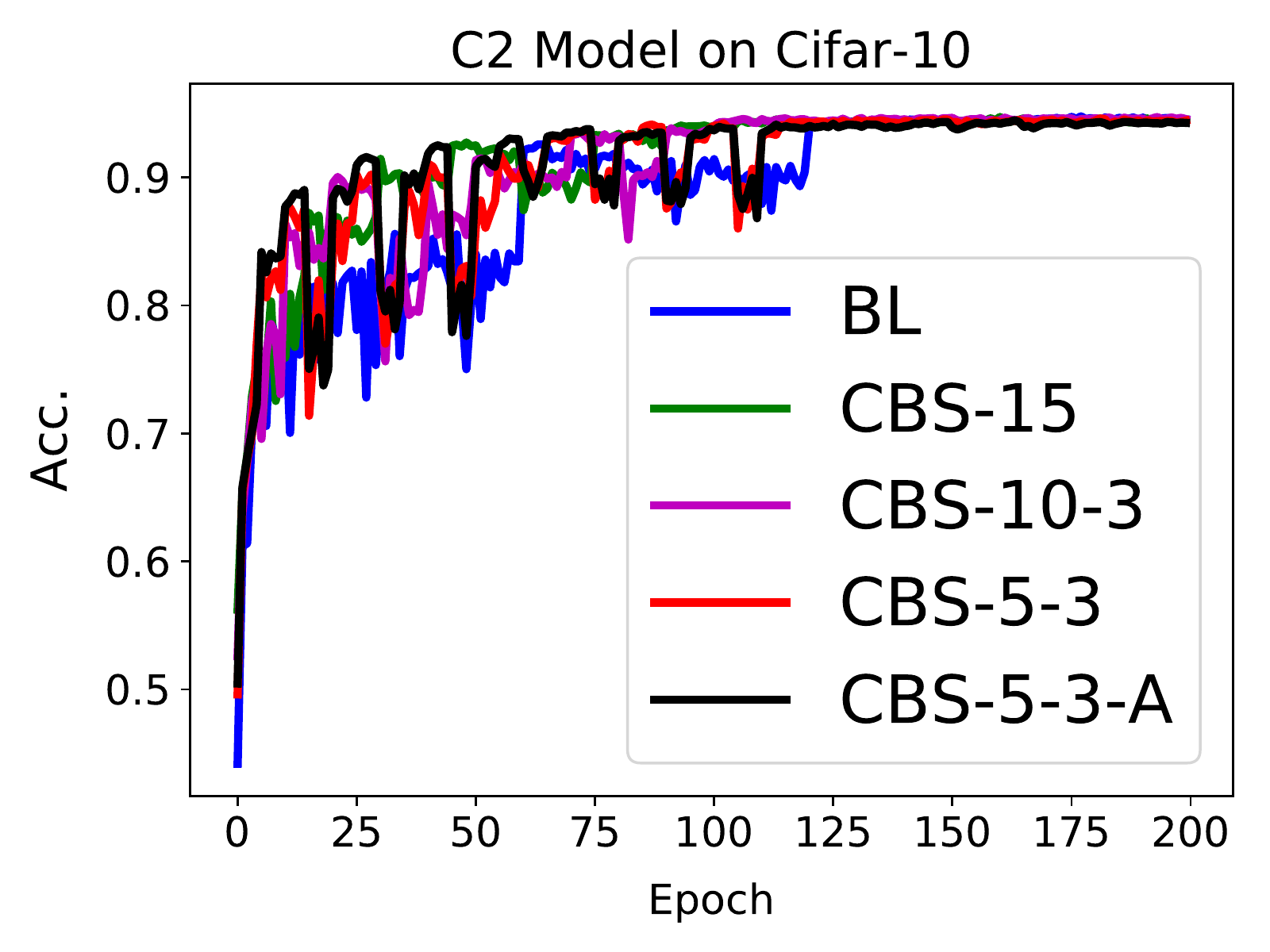}
  \caption{\footnotesize C2 model (WResNet) on Cifar-10. Training set loss (left), and testing set accuracy (right), evaluated as a function of epochs}

  \label{fig:wresnet_cifar10}
\end{figure}

As seen in \fref{fig:wresnet_cifar10}, the training curves of CBS schedules also 
exhibit the aforementioned cyclical spikes both in training loss and testing 
accuracy. Similarly in the previously discussed language experiments, these spikes correspond to cycles in the CBS schedules and can 
be thought of as re-initializations of the neural network weights. We observe that CBS achieves similar performance to the baseline.

\begin{figure}[!htbp]
  \centering
\includegraphics[width=.4\textwidth]{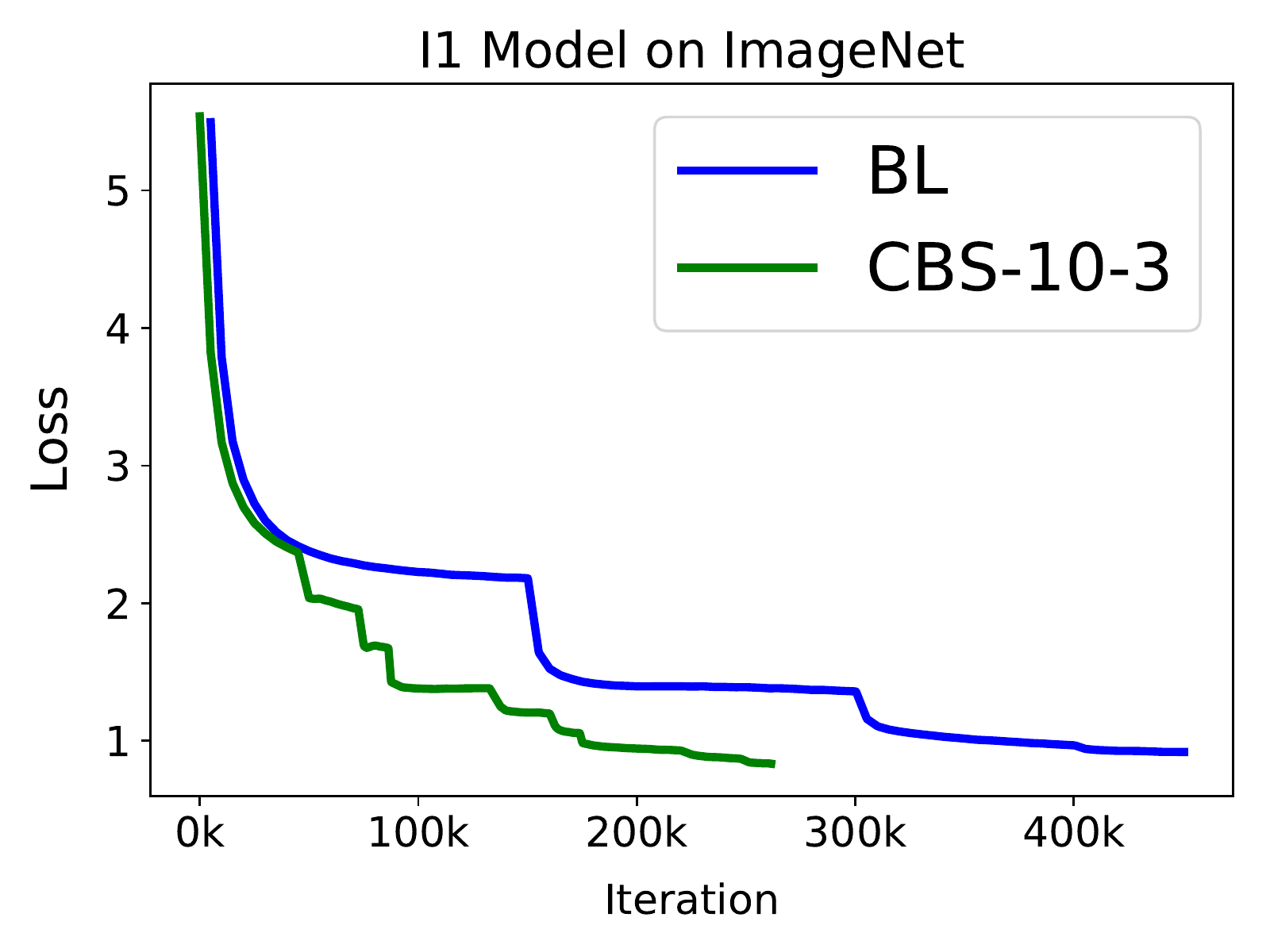}
\includegraphics[width=.4\textwidth]{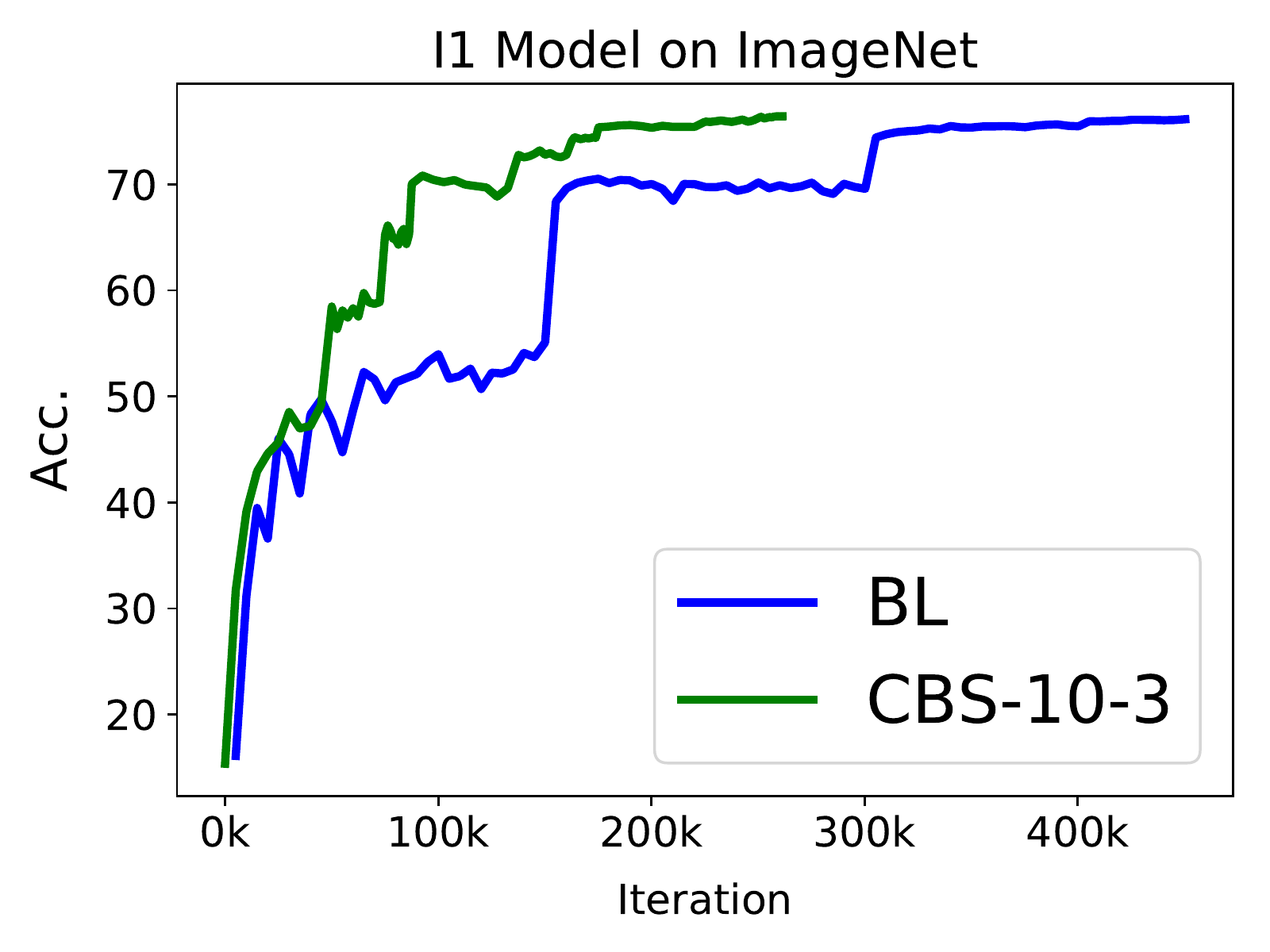}
\includegraphics[width=.4\textwidth]{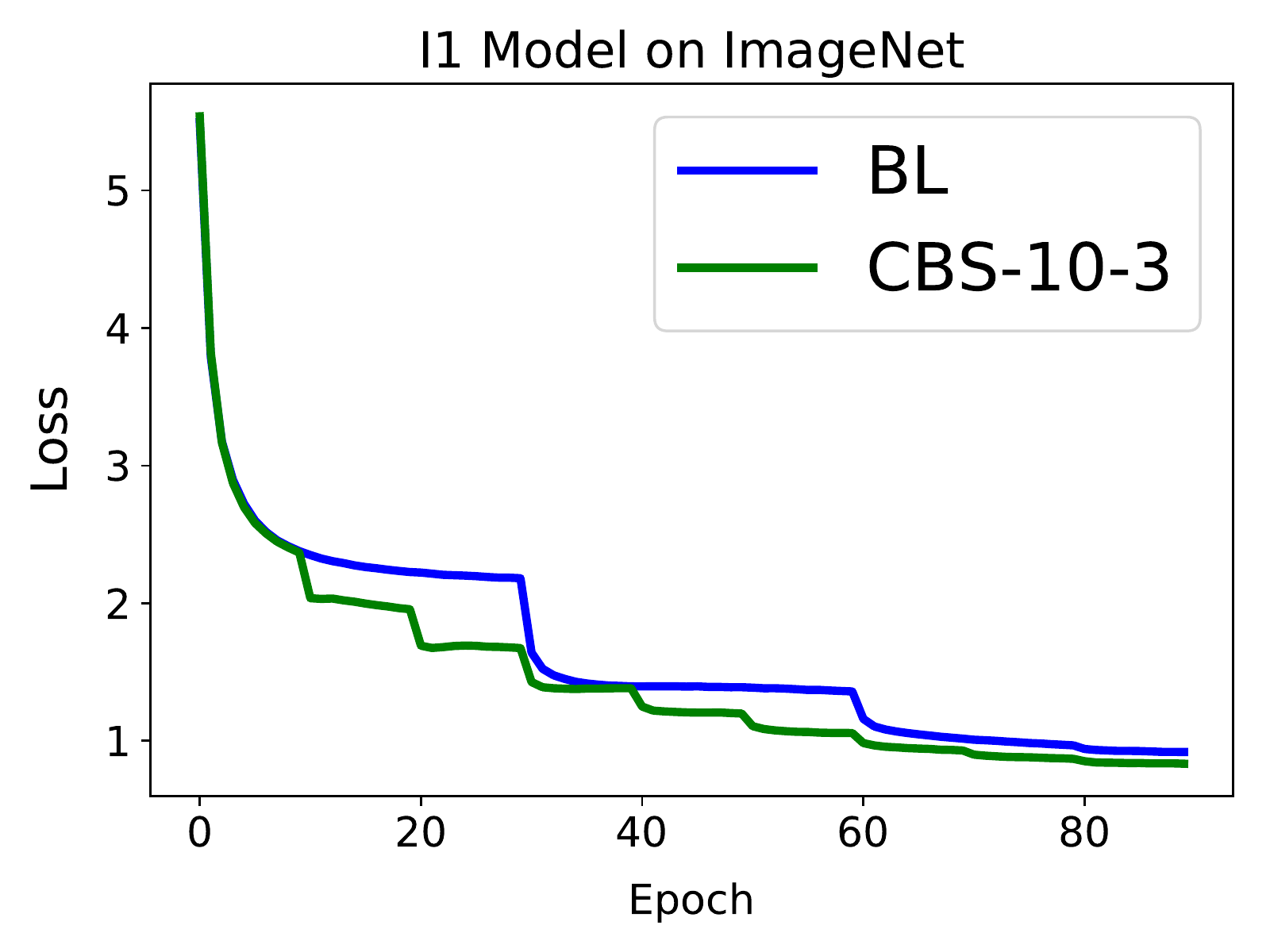}
\includegraphics[width=.4\textwidth]{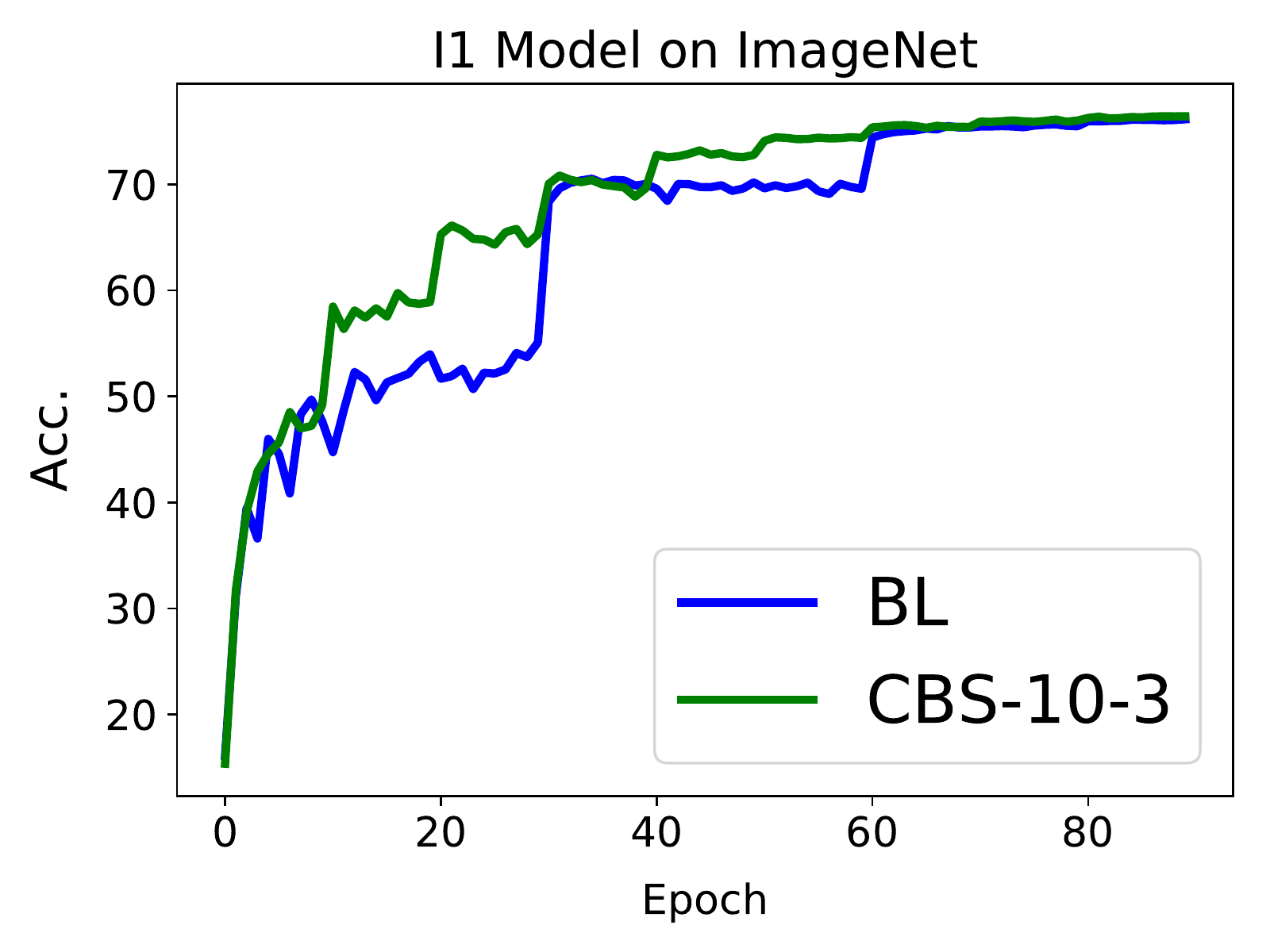}

  \caption{\footnotesize I1 model (ResNet50) on ImageNet. Training set loss (left), and testing set accuracy (right), evaluated as a function of iterations (above) and epochs (below).}
  \label{fig:cbs_imagenet}
\end{figure}

\fref{fig:cbs_imagenet} shows the results of ResNet50 on ImageNet. The baseline trains in  $450k$ 
iterations and reaches $76.134\%$ validation accuracy. With CBS, the final validation accuracy is 
$76.336\%$, trained in $262k$ parameter updates.
CBS outperforms the baseline on both training loss and validation accuracy. 

We offer further support for the hypothesis that CBS schedules are more effective for overfitting neural networks with experiments on model C4, which achieves 94.35\% training 
accuracy 
and 55.55\% testing accuracy on Cifar-10. With CBS-15, we see 90.71\% training 
accuracy 
and 56.44\% testing accuracy, which is a larger improvement than that offered by CBS on convolutional models on Cifar-10.

We also explore combining CBS with the recent adversarial regularization proposed by~\cite{yao2018large}.
Combining CBS-15 on C2 with this strategy improves accuracy to $94.82\%$. This outperforms other schedules shown in  Table~\ref{tab:cbs_cifar10}.
Applying snapshot ensembling on C3 trained with CBS-15-2 leads to improved accuracy of $93.56\%$ as compared to $92.58\%$.
After ensembling ResNet50 on Imagenet with snapshots from the last two cycles, the performance increases to 76.401\% from 75.336\%.

\begin{table}[!htbp]
\caption{\footnotesize Accuracy and number of parameter updates of different models on Cifar-10. The best accuracy and lowest number of iterations are \textbf{bolded}.}
\label{tab:cbs_cifar10}
\centering
\begin{tabular}{lcc|lcc|lcc} \toprule
\multicolumn{3}{c}{AlexNet-like (C1)}  &\multicolumn{3}{c}{WResNet (C2)}  &\multicolumn{3}{c}{ResNet18 (C3)} \\              
\midrule
    {Strategy}                  & {Acc.} & {\# Iters}               &{Strategy}           & {Acc.} & {\# Iters}        &{Strategy}           & {Acc.} & {\# Iters}                     \\
    \midrule
\Gc  Baseline             &86.94            & 35k            & Baseline     &94.53   & 78k                   & Baseline      & \textbf{92.71}  & 63k          \\
\Ga  CBS-10-3          &86.83            & 20k               & CBS-15       &94.46   & 40k                   & CBS-10 & 92.47 & \textbf{32k}         \\
\Gc  CBS-15-2          &86.87            & 26k               & CBS-10-3     &\textbf{94.56}   & 45k          & CBS-5-3  & 92.45 & 37k        \\
\Ga  CBS-5-3           &\textbf{87.03}   & 20k      & CBS-5-3      &94.44   & 45k                   & CBS-15-2 & 92.58 & 48k         \\
\Gc  CBS-5-3-A         &86.75            & \textbf{15k}               & CBS-5-3-A    &94.34   & \textbf{33k}          & CBS-15-2-A & 92.27 & 39k             \\
     \bottomrule 
\end{tabular}
\end{table}

\subsection{Sub-optimal Initialization}\label{sec:bad_init}
Various effective initialization methods~\cite{glorot2010understanding,he2015delving,saxe2013exact,mishkin2015all}
have been proposed previously; however, when presented with new architectures and new tasks,
initialization still needs to be explored empirically and often the final
performance varies greatly with different initializations. In this section, we test if
CBS schedules can alleviate the problem of sub-optimal initialization.

We test a Gaussian initialization with mean $0$ and standard deviation $0.1$ on 
an AlexNet-like model (C1). The baseline (BL) training follows the same setting as described in 
Appendix~\ref{sec:training_outline} and achieves final accuracy $84.27\%$. For CBS, we use cycle width of 10 with 3 steps.
In particular, CBS$_1$ denotes a constant learning rate, and achieves final accuracy $85.41\%$. CBS$_2$ decays the learning rate by a factor of 5 at
epoch 75 and achieves final accuracy $84.95\%$. 
We keep learning rate high during training because a high noise 
level helps $\theta$ escape sub-optimal local minima. 
Notice that all CBS methods achieve better generalization performance than the baseline.
\section{Conclusions}\label{sec:conclusions}
In this work we explored different cyclical batch size (CBS) schedules for training neural networks. We framed the motivation behind CBS schedules through the lens of Bayesian statistical methods, in particular adaptive MCMC algorithms, which seek out better estimates of the
posterior starting with a (poor) prior distribution. In the context of neural network training, this translates to re-initialization of the weights via cycling between large and small batch sizes which control the noise in SGD. We show empirical results which find this cyclical batch size schedule can significantly outperform fixed batch size baselines, especially in networks prone to overfitting or initialized poorly, on the tasks of language modeling, natural language inference, and image classification with LSTMs, CNNs, and ResNets.
In our language modeling experiments, we see that a wide variety of CBS schedules outperform the baseline by up to 7.91 perplexity and up to $33\%$ fewer training iterations. For natural language inference and image classification tasks, we observe a reduction in the number of training iterations of up to $61\%$, which translates directly into reduced runtime. Finally, we demonstrate the flexibility of CBS as a building block for ensembling and adversarial training methods. Ensembling on language modeling yields improvements of up to 11.22 perplexity over the baseline and on image classification, an improvement of up to $1.07\%$ accuracy. Adversarial training in conjunction with CBS gives a bump in image classification accuracy of 0.26\%.

\paragraph{Limitations}
We believe that it is very important for every work to state its limitations (in 
general, but in particular in this area).
We performed an extensive variety of experiments on different tasks in order to comprehensively test the algorithm. 
The primary limitation of our work is that cyclical batch size schedules introduce another
hyper-parameter that requires manual tuning. We note that this is also true
for cyclical learning rate schedules, and hope to address this
using second order methods~\cite{yao2018large} as part of future work.
Furthermore, for well initialized models which are not prone to overfitting, single snapshot
CBS achieves similar performance to the baseline, although the cyclical ensembling provides a modicum of improvement.

\bibliographystyle{plainnat}
\bibliography{ref}

\begin{thebibliography}{28}
\providecommand{\natexlab}[1]{#1}
\providecommand{\url}[1]{\texttt{#1}}
\expandafter\ifx\csname urlstyle\endcsname\relax
  \providecommand{\doi}[1]{doi: #1}\else
  \providecommand{\doi}{doi: \begingroup \urlstyle{rm}\Url}\fi

\bibitem[Bowman et~al.(2015)Bowman, Angeli, Potts, and Manning]{snli:emnlp2015}
Samuel~R. Bowman, Gabor Angeli, Christopher Potts, and Christopher~D. Manning.
\newblock A large annotated corpus for learning natural language inference.
\newblock In \emph{Proceedings of the 2015 Conference on Empirical Methods in
  Natural Language Processing (EMNLP)}. Association for Computational
  Linguistics, 2015.

\bibitem[Chen et~al.(2017)Chen, Zhu, Ling, Wei, Jiang, and
  Inkpen]{chen2017enhanced}
Qian Chen, Xiaodan Zhu, Zhen-Hua Ling, Si~Wei, Hui Jiang, and Diana Inkpen.
\newblock Enhanced lstm for natural language inference.
\newblock In \emph{Proceedings of the 55th Annual Meeting of the Association
  for Computational Linguistics (Volume 1: Long Papers)}, volume~1, pages
  1657--1668, 2017.

\bibitem[Devarakonda et~al.(2017)Devarakonda, Naumov, and
  Garland]{devarakonda2017adabatch}
Aditya Devarakonda, Maxim Naumov, and Michael Garland.
\newblock Adabatch: Adaptive batch sizes for training deep neural networks.
\newblock \emph{arXiv preprint arXiv:1712.02029}, 2017.

\bibitem[Gholami et~al.(2018)Gholami, Azad, Jin, Keutzer, and
  Buluc]{gholami2017integrated}
Amir Gholami, Ariful Azad, Peter Jin, Kurt Keutzer, and Aydin Buluc.
\newblock Integrated model, batch and domain parallelism in training neural
  networks.
\newblock \emph{ACM Symposium on Parallelism in Algorithms and
  Architectures(SPAA'18)}, 2018.

\bibitem[Glorot and Bengio(2010)]{glorot2010understanding}
Xavier Glorot and Yoshua Bengio.
\newblock Understanding the difficulty of training deep feedforward neural
  networks.
\newblock In \emph{Proceedings of the Thirteenth International Conference on
  Artificial Intelligence and Statistics}, pages 249--256, 2010.

\bibitem[Goodfellow et~al.(2014)Goodfellow, Shlens, and
  Szegedy]{goodfellow6572explaining}
Ian~J Goodfellow, Jonathon Shlens, and Christian Szegedy.
\newblock Explaining and harnessing adversarial examples (2014).
\newblock \emph{arXiv preprint arXiv:1412.6572}, 2014.

\bibitem[He et~al.(2015)He, Zhang, Ren, and Sun]{he2015delving}
Kaiming He, Xiangyu Zhang, Shaoqing Ren, and Jian Sun.
\newblock Delving deep into rectifiers: Surpassing human-level performance on
  imagenet classification.
\newblock In \emph{Proceedings of the IEEE International Conference on Computer
  Vision}, pages 1026--1034, 2015.

\bibitem[He et~al.(2016)He, Zhang, Ren, and Sun]{he2016deep}
Kaiming He, Xiangyu Zhang, Shaoqing Ren, and Jian Sun.
\newblock Deep residual learning for image recognition.
\newblock In \emph{Proceedings of the IEEE conference on computer vision and
  pattern recognition}, pages 770--778, 2016.

\bibitem[Hu et~al.(2017)Hu, Yao, and Li]{hu2017adaptive}
Zixi Hu, Zhewei Yao, and Jinglai Li.
\newblock On an adaptive preconditioned crank--nicolson mcmc algorithm for
  infinite dimensional bayesian inference.
\newblock \emph{Journal of Computational Physics}, 332:\penalty0 492--503,
  2017.

\bibitem[Huang et~al.(2017)Huang, Li, Pleiss, Liu, Hopcroft, and
  Weinberger]{huang2017snapshot}
Gao Huang, Yixuan Li, Geoff Pleiss, Zhuang Liu, John~E Hopcroft, and Kilian~Q
  Weinberger.
\newblock Snapshot ensembles: Train 1, get m for free.
\newblock \emph{arXiv preprint arXiv:1704.00109}, 2017.

\bibitem[Jastrz{\k{e}}bski et~al.(2017)Jastrz{\k{e}}bski, Kenton, Arpit,
  Ballas, Fischer, Bengio, and Storkey]{jastrzkebski2017three}
Stanis{\l}aw Jastrz{\k{e}}bski, Zachary Kenton, Devansh Arpit, Nicolas Ballas,
  Asja Fischer, Yoshua Bengio, and Amos Storkey.
\newblock Three factors influencing minima in sgd.
\newblock \emph{arXiv preprint arXiv:1711.04623}, 2017.

\bibitem[Krizhevsky(2014)]{krizhevsky2014one}
Alex Krizhevsky.
\newblock One weird trick for parallelizing convolutional neural networks.
\newblock \emph{arXiv preprint arXiv:1404.5997}, 2014.

\bibitem[Krizhevsky and Hinton(2009)]{krizhevsky2009learning}
Alex Krizhevsky and Geoffrey Hinton.
\newblock Learning multiple layers of features from tiny images.
\newblock Technical report, Citeseer, 2009.

\bibitem[Martin and Mahoney(2017)]{martin2017rethinking}
Charles~H Martin and Michael~W Mahoney.
\newblock Rethinking generalization requires revisiting old ideas: statistical
  mechanics approaches and complex learning behavior.
\newblock \emph{arXiv preprint arXiv:1710.09553}, 2017.

\bibitem[Martin and Mahoney(2018)]{martin2018implicit}
Charles~H Martin and Michael~W Mahoney.
\newblock Implicit self-regularization in deep neural networks: Evidence from
  random matrix theory and implications for learning.
\newblock \emph{arXiv preprint arXiv:1810.01075}, 2018.

\bibitem[Merity et~al.(2016)Merity, Xiong, Bradbury, and
  Socher]{merity2016pointer}
Stephen Merity, Caiming Xiong, James Bradbury, and Richard Socher.
\newblock Pointer sentinel mixture models.
\newblock \emph{arXiv preprint arXiv:1609.07843}, 2016.

\bibitem[Mishkin and Matas(2015)]{mishkin2015all}
Dmytro Mishkin and Jiri Matas.
\newblock All you need is a good init.
\newblock \emph{arXiv preprint arXiv:1511.06422}, 2015.

\bibitem[Roberts and Rosenthal(2009)]{roberts2009examples}
Gareth~O Roberts and Jeffrey~S Rosenthal.
\newblock Examples of adaptive mcmc.
\newblock \emph{Journal of Computational and Graphical Statistics}, 18\penalty0
  (2):\penalty0 349--367, 2009.

\bibitem[Saxe et~al.(2013)Saxe, McClelland, and Ganguli]{saxe2013exact}
Andrew~M Saxe, James~L McClelland, and Surya Ganguli.
\newblock Exact solutions to the nonlinear dynamics of learning in deep linear
  neural networks.
\newblock \emph{arXiv preprint arXiv:1312.6120}, 2013.

\bibitem[Smith(2017)]{smith2017cyclical}
Leslie~N Smith.
\newblock Cyclical learning rates for training neural networks.
\newblock In \emph{Applications of Computer Vision (WACV), 2017 IEEE Winter
  Conference on}, pages 464--472. IEEE, 2017.

\bibitem[Smith and Le(2018)]{smith2018bayesian}
Samuel~L Smith and Quoc~V Le.
\newblock A {Bayesian} perspective on generalization and {Stochastic Gradient
  Descent}.
\newblock \emph{arXiv preprint arXiv:1710.06451}, 2018.

\bibitem[Smith et~al.(2017)Smith, Kindermans, and Le]{smith2017don}
Samuel~L Smith, Pieter-Jan Kindermans, and Quoc~V Le.
\newblock Don't decay the learning rate, increase the batch size.
\newblock \emph{arXiv preprint arXiv:1711.00489}, 2017.

\bibitem[Williams et~al.(2018)Williams, Nangia, and Bowman]{N18-1101}
Adina Williams, Nikita Nangia, and Samuel Bowman.
\newblock A broad-coverage challenge corpus for sentence understanding through
  inference.
\newblock In \emph{Proceedings of the 2018 Conference of the North American
  Chapter of the Association for Computational Linguistics: Human Language
  Technologies, Volume 1 (Long Papers)}, pages 1112--1122. Association for
  Computational Linguistics, 2018.
\newblock URL \url{http://aclweb.org/anthology/N18-1101}.

\bibitem[Xu et~al.(2017)Xu, Roosta-Khorasan, and Mahoney]{xu2017second}
Peng Xu, Farbod Roosta-Khorasan, and Michael~W Mahoney.
\newblock Second-order optimization for non-convex machine learning: An
  empirical study.
\newblock \emph{arXiv preprint arXiv:1708.07827}, 2017.

\bibitem[Yao et~al.(2018{\natexlab{a}})Yao, Gholami, Keutzer, and
  Mahoney]{yao2018large}
Zhewei Yao, Amir Gholami, Kurt Keutzer, and Michael Mahoney.
\newblock Large batch size training of neural networks with adversarial
  training and second-order information.
\newblock \emph{arXiv preprint arXiv:1810.01021}, 2018{\natexlab{a}}.

\bibitem[Yao et~al.(2018{\natexlab{b}})Yao, Gholami, Lei, Keutzer, and
  Mahoney]{yao2018hessian}
Zhewei Yao, Amir Gholami, Qi~Lei, Kurt Keutzer, and Michael~W Mahoney.
\newblock Hessian-based analysis of large batch training and robustness to
  adversaries.
\newblock \emph{arXiv preprint arXiv:1802.08241}, 2018{\natexlab{b}}.

\bibitem[Zagoruyko and Komodakis(2016)]{zagoruyko2016wide}
Sergey Zagoruyko and Nikos Komodakis.
\newblock Wide residual networks.
\newblock \emph{arXiv preprint arXiv:1605.07146}, 2016.

\bibitem[Zaremba et~al.(2014)Zaremba, Sutskever, and
  Vinyals]{zaremba2014recurrent}
Wojciech Zaremba, Ilya Sutskever, and Oriol Vinyals.
\newblock Recurrent neural network regularization.
\newblock \emph{arXiv preprint arXiv:1409.2329}, 2014.

\end{thebibliography}

\clearpage
\appendix
\section{Training Details}\label{sec:training_outline}
Here we catalogue details regarding all tasks, datasets, models, batch schedules, and other hyper-parameters used in our experiments. In all experiments, we try to copy as many hyper-parameters from the original papers as possible.

\textbf{Tasks: }We train networks to perform the following supervised learning tasks:
\begin{itemize}[noitemsep,topsep=0pt,parsep=0pt,partopsep=0pt,leftmargin=*]

\item 
\textbf{Image classification.} 
The network is trained to classify the content of images within a fixed set of object classes.

\item
\textbf{Language modeling.}
The network is trained to predict the last token in a sequence of English words.

\item
\textbf{Natural Language Inference.}
The network is trained to classify the relationship between pairs of English sentences such as that of entailment, contradiction, or neutral.
\end{itemize}

\textbf{Datasets: }We train networks on the following datasets.
\begin{itemize}[noitemsep,topsep=0pt,parsep=0pt,partopsep=0pt,leftmargin=*]

\item 
\textbf{Cifar (image classification).} 
The two Cifar (i.e., Cifar-10/Cifar-100) datasets~\citep{krizhevsky2009learning} contain 50k training images and 10k testing images, and 10/100 label classes. 

\item 
\textbf{ImageNet (image classification).} 
The ILSVRC 2012 classification dataset consists of 1000 label classes, with a total of 1.2 million training images and 50,000 validation images. During training, we crop the image to $224 \times 224$.

\item 
\textbf{PTB (language modeling).} 
The Penn Tree Bank dataset consists of preprocessed and tokenized sentences from the Wall Street Journal. The training set is 929k words, the validation set 73k words, and test set 82k words. The total vocabulary size is 10k, and all words outside the vocabulary are replaced by a placeholder token.

\item 
\textbf{WikiText~2 (language modeling).} 
The Wikitext~2 dataset is modeled after the Penn Tree Bank dataset and consists of preprocessed and tokenized sentences from Wikipedia. The training set is 2089k words, the validation set 218k words, and the test set 246k words. The total vocabulary size is 33k, and all words outside the vocabulary are replaced by a placeholder token.

\item 
\textbf{SNLI (natural language inference).} 
The SNLI dataset \citep{snli:emnlp2015} consists of pairs of sentences annotated with one of three labels regarding textual entailment information: contradiction, neutral, or entailment. The training set contains 550k pairs, and the validation set contains 10k pairs.

\item 
\textbf{MultiNLI (natural language inference.} 
The MultiNLI dataset \citep{N18-1101} is modeled after the SNLI dataset and contains a training set of 393k pairs and a validation set of 20k pairs.
\end{itemize}

\begin{figure}[htbp]
  \centering
\includegraphics[width=.45\textwidth]{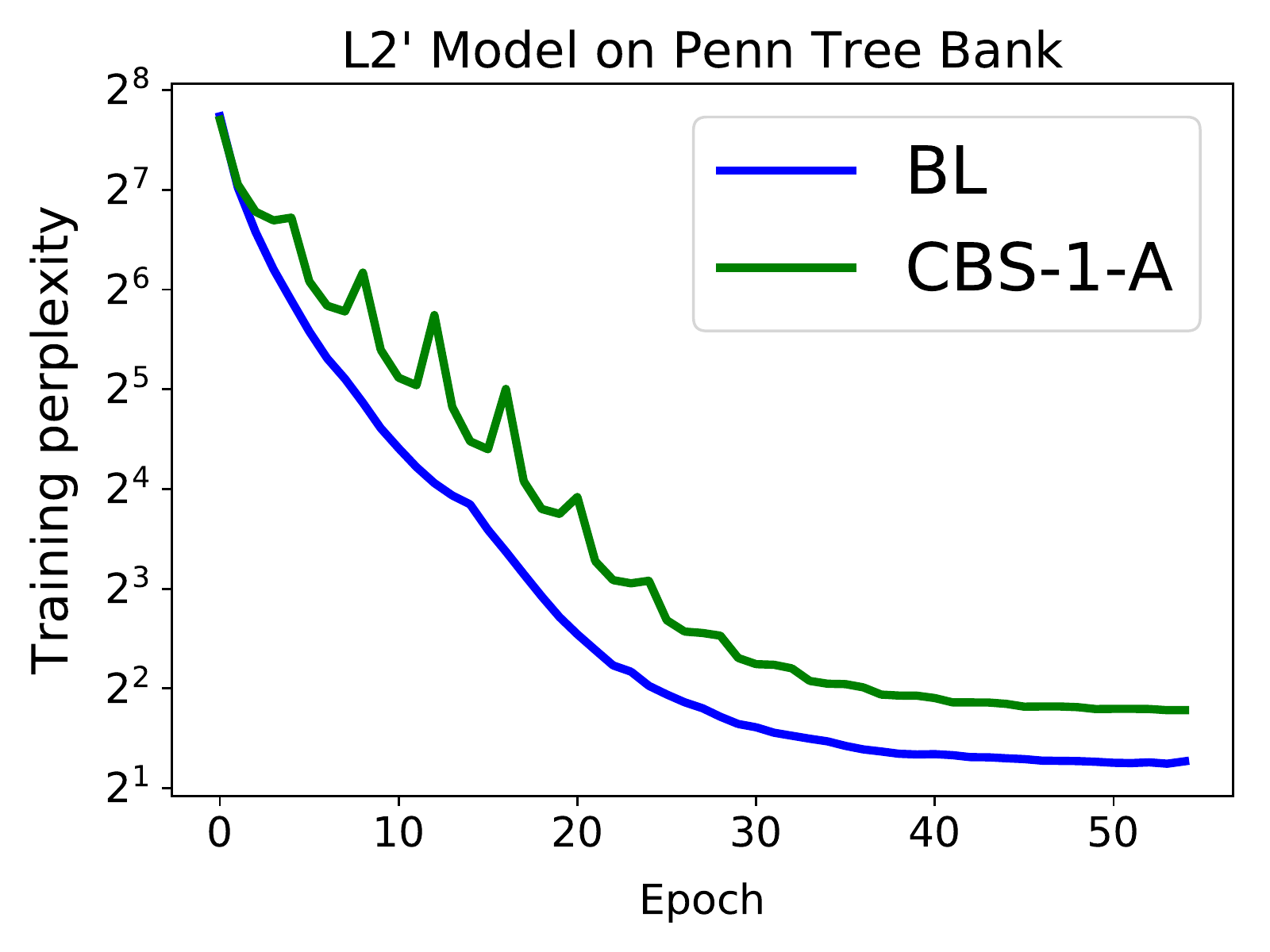}
\includegraphics[width=.45\textwidth]{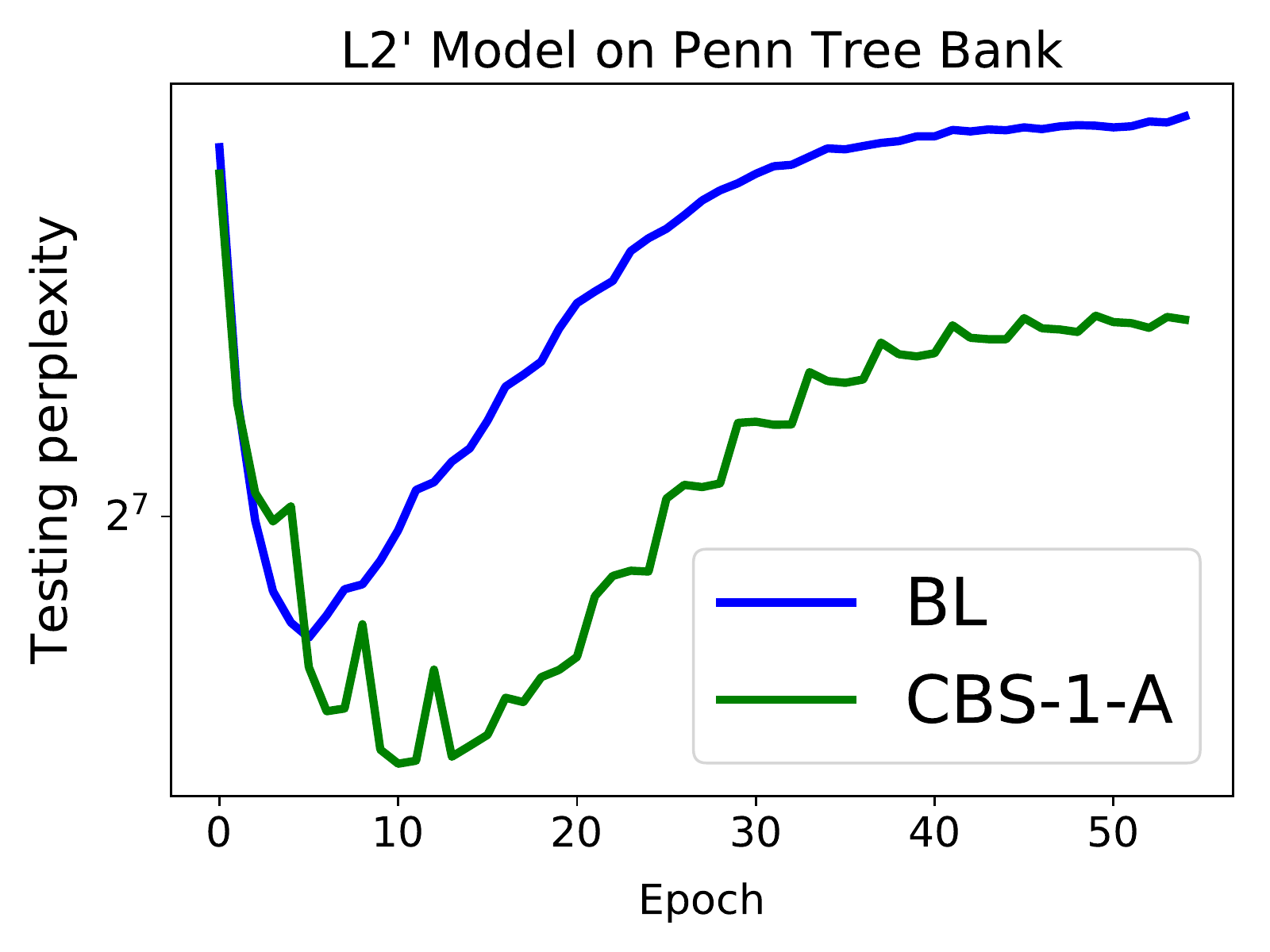}
  \caption{Training (left) and testing (right) perplexity as a function of epoch for overfitting L2' model on Penn Tree Bank.}
  \label{fig:l2'_ptb}
\end{figure}

\textbf{Model Architecture.}
We implement the following neural network architectures. 
\begin{itemize}[noitemsep,topsep=0pt,parsep=0pt,partopsep=0pt,leftmargin=*]

\item \textbf{C1.} AlexNet-like on Cifar-10 dataset as in \citep{yao2018hessian}[C1], trained on the task of image classification. We train for 200 epochs with an initial learning rate $0.02$ which we decay by a factor of 5 at epoch 30, 60. In particular, we use initial learning rate $0.05$ for cyclic scheduling.

\item \textbf{C2.} WResNet 16-4 on Cifar-10 dataset~\citep{zagoruyko2016wide}, trained on the task of image classification. We train for 200 epochs with an initial learning rate $0.1$ which we decay by a factor of 5 at epoch 60, 120, and~180.

\item \textbf{C3.} ResNet20 on Cifar-10 dataset~\citep{he2016deep}. We train it for 160 epochs with initial learning rate $0.1$, and decay a factor of 5 at epoch 80, 120. In particular, we use initial learning rate $0.05$ for cyclic scheduling.

\item \textbf{C4.} MLP3 network from~\citep{martin2018implicit}. The network consists of 3 fully connected layers with 512 units each and ReLU activations. As a baseline, we train this network with vanilla SGD for 240 epochs with a batch size of 100 and an initial learning rate of 0.1, which is decayed by a factor of 10 at  150 and 225 epochs.

\item \textbf{I1.} ResNet50 on ImageNet dataset~\citep{he2016deep}, trained on the task of image classification for 90 epochs with initial learning rate $0.1$ which we decay by a factor of 10 at epoch 30, 60 and 80.

\item \textbf{L1.} Medium Regularized LSTM~\citep{zaremba2014recurrent}, trained on the task of language modeling. We use 50\% dropout on non-recurrent connections and train for 39 epochs with initial learning rate of 20, decaying by a factor of 1.2 every epoch after epoch 6. We set a backpropagation-through-time limit of 35 steps and clip the max gradient norm at 0.25.

\item \textbf{L2.} Large Regularized LSTM~\citep{zaremba2014recurrent}, trained on the task of language modeling. We use 65\% dropout on non-recurrent connections and train for 55 epochs with initial learning rate of 20, decaying by a factor of 1.15 every epoch after epoch 14. We set a backpropagation-through-time limit of 35 steps and clip the max gradient norm at 0.5.

\item \textbf{L1', L2'} Identical to L1, L2 except for lower dropout: 0.2, 0.3 respectively. Leads to significant overfitting, evidenced by test perplexity curve in \fref{fig:l2'_ptb}.

\item \textbf{E1.} ESIM~\citep{chen2017enhanced}. We train the base ESIM model without the tree-LSTM, as in ~\citep{N18-1101}, on the task of natural language inference with ADAM for 10 epochs on MultiNLI and also SNLI.
\end{itemize}

\textbf{Training Schedules:}
We use the following batch size schedules
\begin{itemize}[noitemsep,topsep=0pt,parsep=0pt,partopsep=0pt,leftmargin=*]
\item \textbf{BL.} Use a fixed small batch size as specified in the original paper introducing the model or as is~standard. 
\item \textbf{CBS-$k$(-$n$).} Use a Cyclical Batch Size schedule, where $k$ is the width of each step measured in epochs and $n$ is the integer number of steps per cycle. When $n$ is not specified it refers to the default value of 4. At the beginning of each cycle the batch size is initialized to the base batch size, and after each step it is then doubled.
\item \textbf{CBS-$k$(-$n$)-A}. Use an aggressive Cyclical Batch Size schedule, which is equivalent to the original CBS schedule except after every step the batch size is quadrupled.
\item \textbf{CBS-$k$(-$n$)-T}. Use a triangular Cyclical Batch Size schedule, which is modeled after the triangular schedule. Each cycle consists of $n$ steps doubling the batch size after each step, then $n-2$ symmetrical steps halving the batch size after each step.
\end{itemize}

In all language modeling CBS experiments, we use an initial batch size of 10, that is, 
half the baseline batch size as reported in the respective papers of each baseline
model tested. The intuition behind starting with a smaller batch size is to introduce 
additional noise to help models escape sub-optimal local minima. 

For adversarial training used in image classification, we use FGSM method~\cite{goodfellow6572explaining} to generate adversarial examples. Adversarial training is implemented for the first half training epochs.

\section{Additional Results}
\label{sxn:app:additional results}
This section shows additional experiment results.

\begin{table}[!htbp]
\caption{\footnotesize Test perplexity of L1, L2 snapshot ensembled models on Penn Tree Bank (PTB) and WikiText-2 (WT2) datasets. Each CBS ensemble model is trained on its best-performing CBS schedule and each baseline ensemble model is snapshotted at the same epochs as its corresponding CBS ensemble model.}
\label{tab:ensemble-results}
\centering
\begin{tabular}{lcc|cc} \toprule
  & {L1 PTB}    & {L1 WT2}   & {L2 PTB} & {L2 WT2}    \\              
\midrule
\Ga BL Single   &   83.13   &   96.41   &   79.34   &   99.69    \\
\Gc	BL Ens.	    &	82.22	&	97.18	&	76.52	&	89.99	\\
\Ga	CBS	Ens     &	81.51	&	94.27   &	76.14   &   88.47   \\
\bottomrule 
\end{tabular}
\end{table}

\begin{table}[!htbp]
\caption{\footnotesize Final testing perplexity and best testing perplexity of low-dropout L1' and L2' models on Penn 
Tree Bank (PTB) and WikiText~2 (WT2) datasets. The best perplexity values are \textbf{bolded}.}
\label{tab:lm-results-overfit}
\centering
\begin{tabular}{lcc|cc|cc|cc} \toprule
  &\multicolumn{2}{c}{L1' on PTB}       &\multicolumn{2}{c}{L1' on WT2}    &\multicolumn{2}{c}{L2' on PTB} &\multicolumn{2}{c}{L2' on WT2}\\              
\midrule
Schedule    & {Final}    & {Best}    & {Final}    & {Best}    & {Final}    & {Best}    & {Final}    & {Best}      \\
\midrule
\Gc BL & 132.4	&	108.4	&	132.2	&	129.8	&	231.7	&	106.9	&	177.3	&	136.6 \\
\Ga CBS-10-A & 119.4	&	105.4	&	119.0	&	115.7	&	177.2	&	97.8	&	\textbf{145.1}	&	116.7 \\
\Gc CBS-5-A & 115.1	&	95.6	&	116.6	&	111.4	&	257.0	&	\textbf{88.3}	&	178.5	&	106.9 \\
\Ga CBS-1-A & \textbf{106.4}	&	\textbf{93.6}	&	\textbf{113.5}	&	\textbf{104.5}	&	\textbf{171.4}	&	88.7	&	147.1	&	\textbf{100.4} \\
\bottomrule 
\end{tabular}
\end{table}

\begin{figure}[!htbp]
  \centering
\includegraphics[width=.45\textwidth]{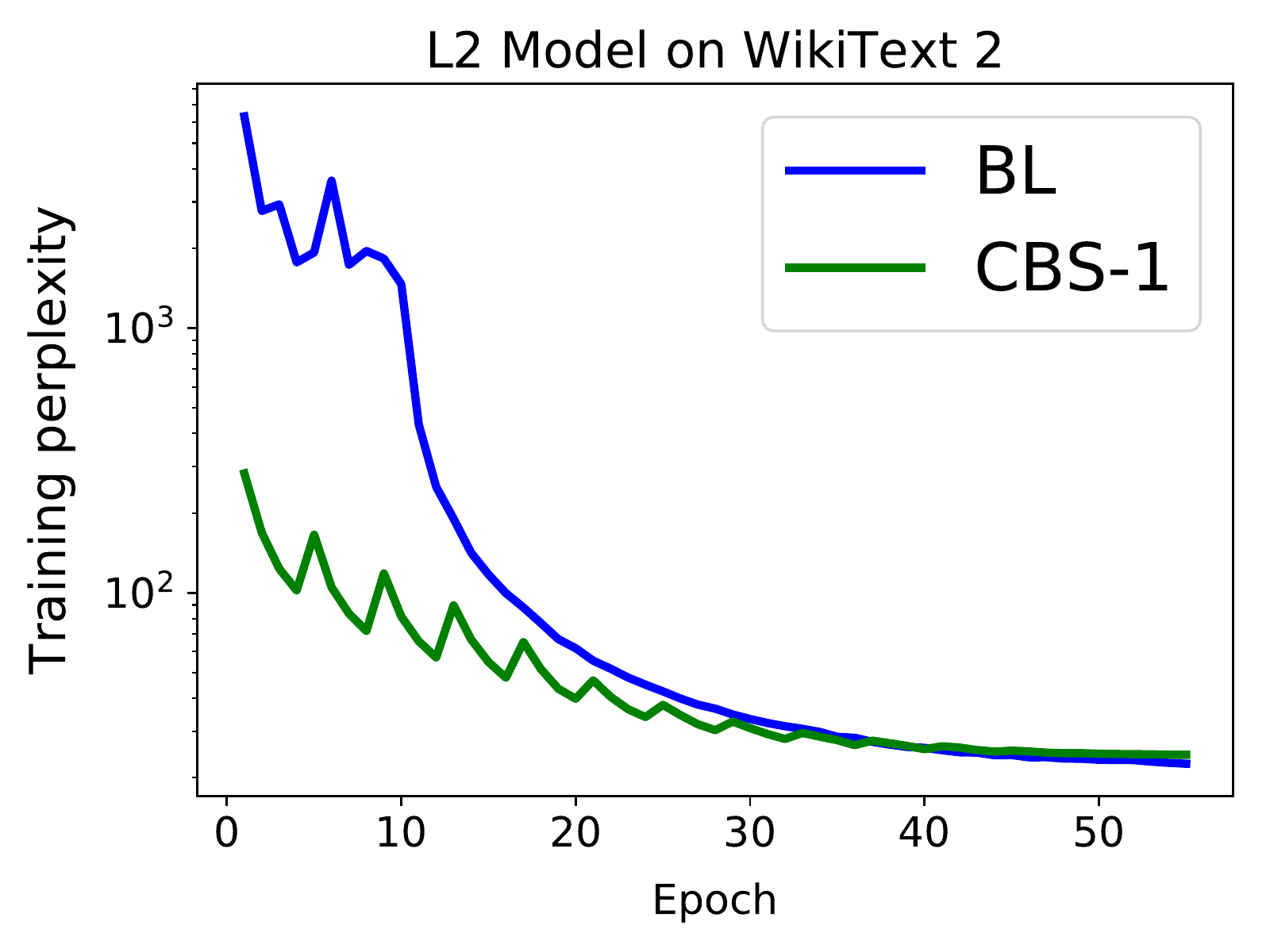}
\includegraphics[width=.45\textwidth]{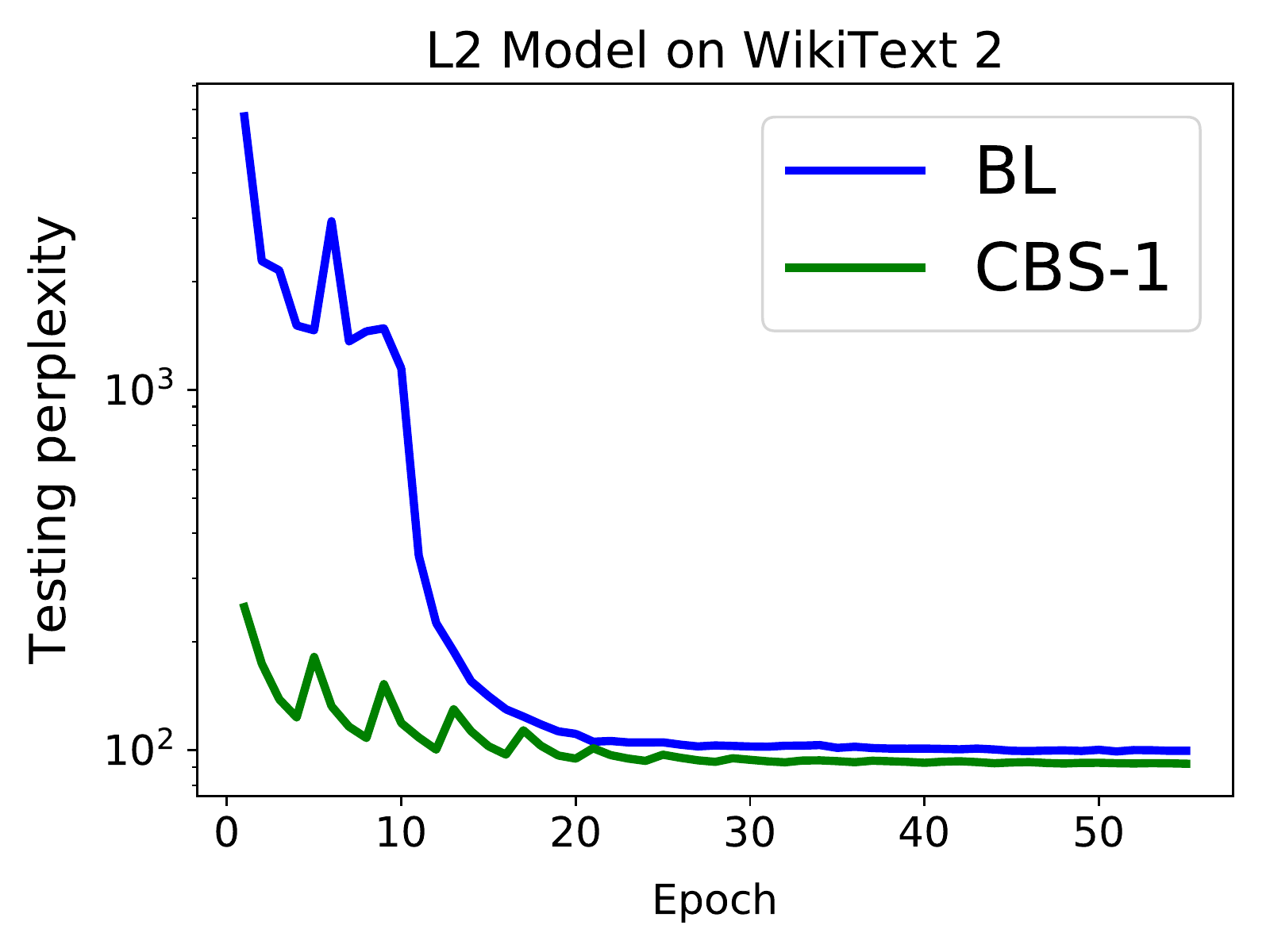}
  \caption{Training (left) and testing (right) perplexity as a function of epoch for L2 model on WikiText~2.}
  \label{fig:l2_wiki}
\end{figure}

\end{document}